%% file: main.tex
\documentclass[conference]{IEEEtran}
\IEEEoverridecommandlockouts

\input{packages}

\begin{document}

\title{FedCCL: Federated Clustered Continual Learning Framework for Privacy-focused Energy Forecasting}

\author{\IEEEauthorblockN{Michael A. Helcig}
\IEEEauthorblockA{\textit{Distributed Systems Group} \\
\textit{TU Wien}\\ 
michael.helcig@student.tuwien.ac.at}
\and
\IEEEauthorblockN{Stefan Nastic}
\IEEEauthorblockA{\textit{Distributed Systems Group} \\
\textit{TU Wien}\\
snastic@dsg.tuwien.ac.at}
}

\maketitle

\input{abstract}
\input{keywords}
\input{sections/introduction}
\input{sections/framework}
\input{sections/case_study}
\input{sections/evaluation}

\input{sections/related_work}
\input{sections/conclusion}
\input{acknowledgements}

\bibliographystyle{IEEEtran}
\bibliography{main}

\end{document}

%% file: packages.tex
\usepackage{cite}
\usepackage{float}
\usepackage{amsmath,amssymb,amsfonts}
\usepackage{algorithmic}
\usepackage[pdftex]{graphicx}
\usepackage{textcomp}
\usepackage{xcolor}
\usepackage{algorithmic}
\usepackage{booktabs}
\usepackage{lipsum}
\usepackage{stfloats} 
\usepackage{amsmath}
\usepackage{hyperref}
\usepackage[ruled]{algorithm2e}

\def\BibTeX{{\rm B\kern-.05em{\sc i\kern-.025em b}\kern-.08em
    T\kern-.1667em\lower.7ex\hbox{E}\kern-.125emX}}

%% file: abstract.tex
\begin{abstract}
Privacy-preserving distributed model training is crucial for modern machine learning applications, yet existing Federated Learning approaches struggle with heterogeneous data distributions and varying computational capabilities. Traditional solutions either treat all participants uniformly or require costly dynamic clustering during training, leading to reduced efficiency and delayed model specialization. We present FedCCL (Federated Clustered Continual Learning), a framework specifically designed for environments with static organizational characteristics but dynamic client availability. By combining static pre-training clustering with an adapted asynchronous FedAvg algorithm, FedCCL enables new clients to immediately profit from specialized models without prior exposure to their data distribution, while maintaining reduced coordination overhead and resilience to client disconnections. Our approach implements an asynchronous Federated Learning protocol with a three-tier model topology — global, cluster-specific, and local models — that efficiently manages knowledge sharing across heterogeneous participants. Evaluation using photovoltaic installations across central Europe demonstrates that FedCCL's location-based clustering achieves an energy prediction error of 3.93\% (±0.21\%), while maintaining data privacy and showing that the framework maintains stability for population-independent deployments, with 0.14 percentage point degradation in performance for new installations. The results demonstrate that FedCCL offers an effective framework for privacy-preserving distributed learning, maintaining high accuracy and adaptability even with dynamic participant populations.
\end{abstract}

%% file: keywords.tex

%% file: sections/introduction.tex
\section{Introduction} \label{section:introduction}
The Federated Learning (FL) paradigm \cite{yang_federated_2019, konecny_federated_2016} has emerged as a pivotal solution for privacy-preserving machine learning, enabling multiple participants to collaboratively train models while maintaining data privacy. This approach has found significant applications across domains where sensitive information must remain decentralized, such as healthcare, finance, and industrial systems. Traditional FL frameworks operate through synchronous rounds of communication, where participants collectively train a single shared model. However, this approach often proves inefficient and inflexible, particularly when dealing with heterogeneous data distributions and varying computational capabilities across participants~\cite{xu_asynchronous_2023, gajanin2024HARFederatedLearning}.

Current challenges in FL deployments include balancing model specialization, meaning the ability to excel at specific client scenarios, with overall generalization, which ensures robust performance across diverse client populations. While many environments have stable organizational characteristics (such as geographic location or hardware specifications), they face dynamic challenges in client availability and connectivity, another issue currently explored~\cite{mammen_federated_2021, bharati_federated_2022}. Existing solutions typically address static or dynamic aspects independently, but not both. 

In terms of communication, synchronous FL approaches~\cite{h_brendan_mcmahan_communication-efficient_2016} require all participants to complete their local training before model aggregation, leading to potential bottlenecks and reduced system efficiency when participants operate at different speeds or face connectivity issues. Asynchronous FL algorithms \cite{cong_xie_asynchronous_2019, xu_asynchronous_2023, chen_asynchronous_2020} address these limitations by enabling independent model updates. The challenge of non-iid (non-independent and identically distributed) data has sparked significant research in Clustered Federated Learning (CFL) approaches. Recent frameworks like Iterative Clustered FL (ICFL)\cite{yan_clustered_2024} propose dynamic clustering mechanisms based on model behavior during training. Weight-driven approaches \cite{md_sirajul_islam_fedclust_2024} exploit correlations between local model weights and data distributions to form clusters. However, these methods typically require multiple training rounds to establish stable clusters and may not immediately benefit new participants joining the federation.

Previous research has extensively explored both solar prediction methodologies \cite{kumari_deep_2021, qing_hourly_2018, yu_deep_2024} as well as the application of FL. Approaches range from hierarchical clustering for residential load forecasting \cite{christopher_briggs_federated_2022} to fuzzy clustering for solar power generation \cite{yoo_fuzzy_2022}. Recent work has explored deep learning models for PV power forecasting \cite{yu_deep_2024}, incorporating domain knowledge \cite{luo_deep_2021} and simplified LSTM \cite{hochreiter_long_1997} architectures \cite{liu_simplified_2021}. There is also some work concerning pre-training clustering \cite{tun_federated_2021}, but without a focus on transfer capabilities, asynchronous rounds and production forecasting. Load forecasting was addressed several times. \cite{saputra_energy_2019, tun_federated_2021, richter_advancing_2024}

This paper presents FedCCL (Federated Clustered Continual Learning), a framework that addresses these challenges through a combination of pre-training clustering and asynchronous Federated Learning. Unlike most of the existing approaches that perform clustering during or after training \cite{felix_sattler_clustered_2020, ghosh_efficient_2022}, FedCCL employs DBSCAN clustering based on static characteristics before training begins. This approach enables immediate model specialization while reducing coordination overhead. Furthermore, participants can belong to multiple clusters simultaneously, facilitating more nuanced knowledge sharing than strict partitioning approaches \cite{yoo_fuzzy_2022}. This paper's main contributions include:

\begin{itemize}
    \item FedCCL Framework: A Federated Learning framework that integrates clustered pre-training with an enhanced asynchronous FedAvg algorithm. The framework operates through a two-phase approach, initially clustering clients based on their inherent system properties before training, followed by client-driven updates with model locking during aggregation. Mitigating the performance degradation typically seen in asynchronous Federated Learning with heterogeneous data while maintaining reduced overhead. 
    \item FedCCL Predict \& Evolve: Through our system property-based clustering approach, FedCCL creates a framework that provides a specialized model for newly joining clients without requiring prior exposure to their specific data distributions. In the "Predict" phase, new clients can immediately benefit from these highly specialized models to generate predictions. As clients begin contributing their own data, they enter the "Evolve" phase, where they participate in training and refining cluster-specific models. Our evaluation demonstrates this capability through robust generalization metrics, where models achieve nearly identical performance levels for both training and independent populations, with mean error rates showing minimal degradation of only 0.14 percentage points for new installations.
    \item FedCCL Case Study: We validate our framework through implementation in solar production forecasting, using real-world data from sites across central Europe, clustered by geographical location and panel orientation. Our experimental results demonstrate superior accuracy compared to both traditional asynchronous Federated Learning and centralized approaches, confirming its effectiveness. In our solar production forecasting implementation using real-world data, the framework with location-based clustering achieved mean errors of 6.44\% ($\pm$0.17\%) for power prediction and 3.93\% ($\pm$0.21\%) for energy prediction, demonstrating superior accuracy compared to the centralized baseline which had higher errors of 6.73\% ($\pm$0.14\%) and 4.18\% ($\pm$0.18\%) respectively.
\end{itemize}

This paper is organized into six sections. We begin by introducing the problem and motivation (Section \ref{section:introduction}), followed by a detailed description of the FedCCL framework and its clustered continual learning approach (Section \ref{section:framework}). We then present an energy forecasting case study (Section \ref{section:case_study}) with experimental results that validate our methodology (Section \ref{section:evaluation}). The paper concludes with a review of related work (Section \ref{section:related_work}) and a discussion of future research directions. (Section \ref{section:conclusion})

%% file: sections/framework.tex
\newpage
\section{FedCCL Framework} \label{section:framework}
The proposed FedCCL framework is designed as a centralized and horizontal system \cite{yang_federated_2019}, it uses a central server to coordinate learning while focusing on scenarios where clients share the same feature space. This section outlines the framework's architecture and operational workflow.

\subsection{Architecture Overview}
\begin{figure}[htbp]
\centerline{\includegraphics[width=0.4\textwidth]{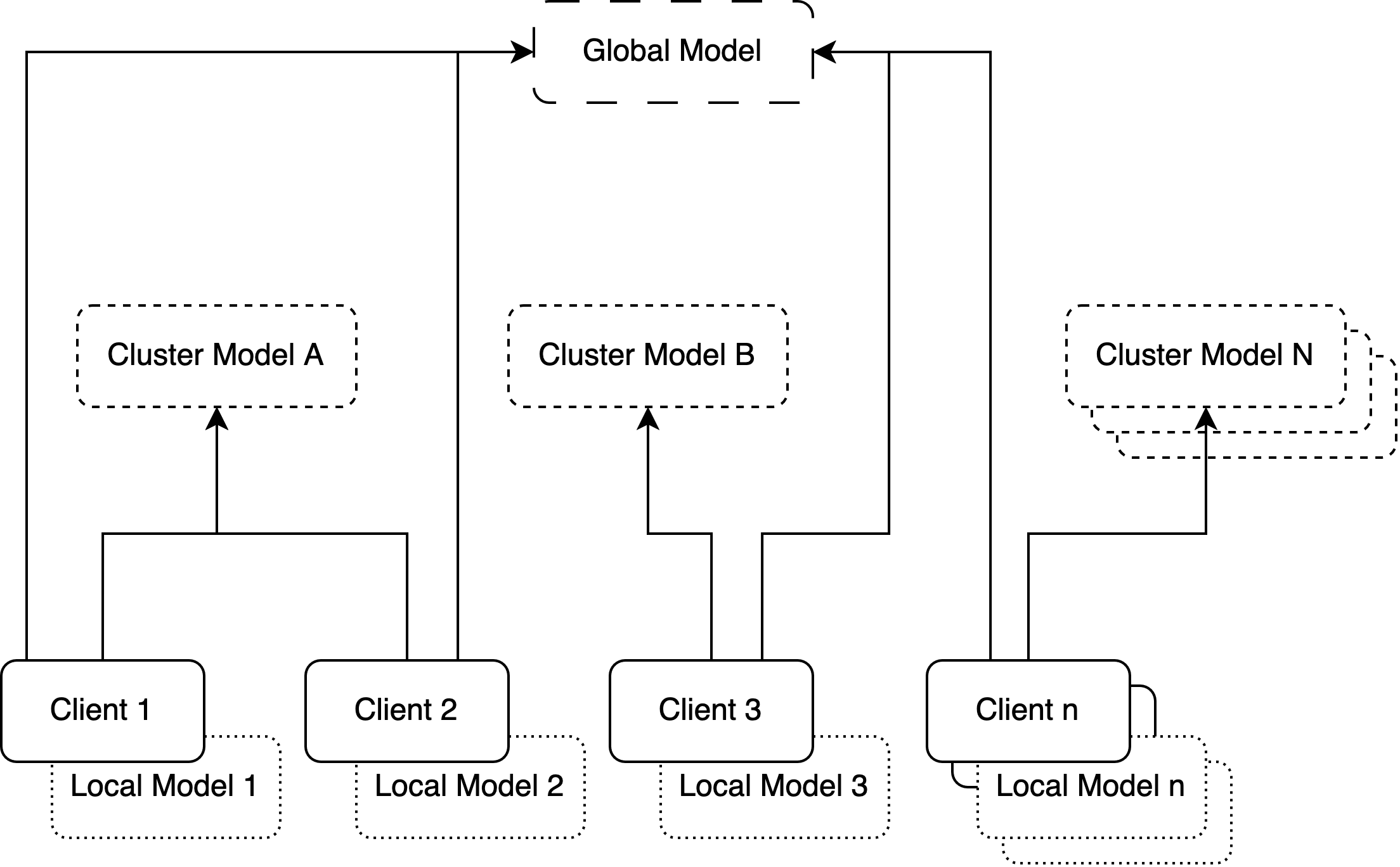}}
\caption{Model Hierarchy of FedCCL}
\label{fig:model_hierarchy}
\end{figure}

FedCCL operates on a three-tier model hierarchy (as illustrated in Figure \ref{fig:model_hierarchy}) that balances global knowledge sharing with specialized learning. At the highest level, a global model (sparsely-dashed) synthesizes knowledge across all clients, capturing common patterns and general features applicable to the entire system. The middle tier consists of cluster-specific models (dashed), each specialized for a group of similar clients. These cluster models, labeled as Cluster Model A through N, refine the global knowledge to better serve the particular characteristics and requirements of their respective client groups, enabling more targeted and efficient learning. At the lowest tier, local models on individual clients (dotted) provide fine-grained specialization, allowing each participant (Client 1 through Client n) to adapt to their specific needs while maintaining the privacy of their local data. This hierarchical structure ensures that the global model captures broad trends, cluster models address group-specific patterns, and local models fine-tune these insights to excel in client-specific tasks, thereby mitigating challenges posed by non-iid data.

\subsection{Pre-training Clustering}
One of FedCCL's distinctive features is its pre-training clustering phase, which precedes model training. Unlike traditional FL approaches, FedCCL employs DBSCAN \cite{martin_ester_density-based_1996} to establish a stable cluster structure based on invariant client characteristics such as geographic location, hardware specifications, or operational parameters. This density-based approach naturally identifies optimal client groupings while filtering outliers. The framework can be extended with the incremental DBSCAN variant \cite{goos_incremental_1998}, enabling network expansion without disrupting established structures. This Predict \& Evolve capability of FedCCL allows new clients to immediately benefit from specialized models tailored to their cluster's characteristics without requiring complete reclustering. By establishing these relationships before training, FedCCL creates a foundation for targeted model specialization while minimizing computational overhead typically associated with dynamic clustering approaches.

\subsection{Asynchronous Training Process}
The core of FedCCL implements a novel asynchronous FL approach, as detailed in Algorithm \ref{alg:training_process}. After initialization (lines 1-2), each client operates independently and in parallel (line 3). During training, clients first update their local models using their private datasets (lines 5-6). Following the local update, they participate in parallel cluster-specific training sessions (lines 7-11), where they request the current cluster model state, perform local training, compute update deltas, and submit their contributions. Finally, clients engage with the global model in a similar manner (lines 12-15). The server handles these asynchronous updates through a dedicated handler function (lines 19-25) that locks the specific model during aggregation to ensure consistency while maximizing parallel operations. In contrast to synchronous FL, this design removes barriers, making it particularly suitable for settings with varying client network conditions.

\begin{algorithm}[ht]
\label{alg:training_process}
\small
\LinesNumbered
\SetAlgoLined
\KwData{$K$ clients with local datasets $D_k$}
\KwResult{Trained models across hierarchy levels}

\BlankLine
Initialize server model store with global and cluster models\;
Initialize local model an set of cluster keys $client\_clusters$ for each client\;

\BlankLine
\For{each \textbf{distributed} client $k$ \textbf{in parallel}}{
    \While{training}{
        Preprocess local data $D_k$\;
        \tcp{Train local model}
        $w_{k_{local}} \leftarrow$ TrainModel($w_{local}$, $D_k$)\;
        \tcp{Train cluster models}
        \For{cluster\_key in client\_clusters \textbf{in parallel}}{
            $w \leftarrow$ RequestModel(level = cluster, cluster\_key)\;
            $w_k \leftarrow$ TrainModel($w$, $D_k$)\;
            $\delta_k \leftarrow$ ComputeModelMetaDelta()\;
            HandleModelUpdate(level = cluster, cluster\_key = cluster\_key, $w_{updated} = w_k$, $\delta_{new} = \delta_k$)\;
        }
        \tcp{Train global model}
        $w \leftarrow$ RequestModel(level = global)\;
        $w_k \leftarrow$ TrainModel($w$, $D_k$)\;
        $\delta_k \leftarrow$ ComputeModelDelta()\;
        HandleModelUpdate(level = 'global', $w_{updated} = w_k$, $\delta_{new} = \delta_k$)\;
    }
}
\BlankLine
\tcp{Server Update Handler}
\SetKwFunction{FHandleUpdate}{HandleModelUpdate}
\SetKwProg{Fn}{Function}{:}{}
\Fn{\FHandleUpdate{level, cluster\_key, $w_{updated}$, $\delta_{new}$}}{
    $m \leftarrow$ GetModel(level, cluster\_key)\;
    \If{AcquireLock(m)}{
        $m \leftarrow$ AggregateModels($m$, $w_{updated}$, $\delta_{new}$)\;
        UpdateModelStore(level, cluster\_key, $m$)\;
        ReleaseLock(m)\;
    }
}
\caption{FedCCL Training Process}
\end{algorithm}

\newpage

\subsection{Model Aggregation Strategy}
FedCCL builds upon FedAvg's \cite{h_brendan_mcmahan_communication-efficient_2016} weighted averaging approach detailed in Algorithm \ref{alg:model_aggregation}. The aggregation process, performed entirely on the server, first evaluates whether the model has been updated by another client by checking if the difference in rounds is exactly 1 (line 1). If this condition is met, it returns the updated model model without aggregation (line 2). When updates are non-sequential, the algorithm computes the total number of training samples (line 4) and performs a layer-wise weighted aggregation (lines 7-11). Each client's influence is proportional to the number of new samples learned in that training round, relative to the total samples processed, determined through ratios calculated for both the base model and new updates. The framework then updates essential metadata (lines 11-13), including cumulative sample count, training epochs, and federation rounds, before returning the aggregated model (line 14). This metadata enables systematic tracking of the federation's learning progress and ensures proper model versioning.

\begin{algorithm}[ht]
\label{alg:model_aggregation}
\small
\LinesNumbered
\SetAlgoLined
\SetKwFunction{FAggregateModels}{AggregateModels}
\SetKwProg{Fn}{Function}{:}{}
\Fn{\FAggregateModels{$w_{base}$, $w_{updated}$, $\delta_{new}$}}{
    \tcp{Check if updates are sequential (i.e round of updated model is exactly one ahead of current)}
    \If{$w_{updated}.\text{round} == w_{base}.\text{round} + 1$}{
        \Return $w_{updated}$\;
    }
    
    $samples\_total \leftarrow w_{base}.samples\_learned + w_{updated}.samples\_learned$\;
    
    Initialize $w_{aggregated}$ with zeros\;
    
    \For{each weight layer $i$}{
        $ratio_{base} \leftarrow \frac{w_{base}.samples\_learned}{samples\_total}$\;
        $ratio_{new} \leftarrow \frac{w_{updated}.samples\_learned}{samples\_total}$\;
        
        $w_{aggregated}[i] \leftarrow w_{base}[i] \times ratio_{base} + w_{updated}[i] \times ratio_{new}$\;
    }
    
    \tcp{Update model metadata}
    $meta.samples\_learned \leftarrow w_{base}.samples\_learned + \delta_{new}.samples\_learned$\;
    $meta.epochs\_learned \leftarrow w_{base}.epochs\_learned + \delta_{new}.epochs\_learned$\;
    $meta.round \leftarrow w_{base}.round + \delta_{new}.round$\;
    
    \Return ModelData($meta$, $w_{aggregated}$)\;
}
\caption{FedCCL Model Aggregation}
\end{algorithm}

\subsection{Continual Learning and Convergence}
FedCCL implements an asynchronous and continuous learning approach where clients can join or leave the training process at any time, and the system can theoretically continue learning indefinitely. Unlike traditional Federated Learning systems that operate in synchronized rounds, FedCCL's asynchronous nature allows clients to independently retrieve the current model state, perform local training, and contribute updates back to the system at their own pace. This flexibility eliminates the need for coordination between clients and enables continuous model improvement without predetermined stopping points. FedCCL addresses the challenge of catastrophic forgetting using a regularization-based approach \cite{kirkpatrick_overcoming_2017}. This method, often referred to as L2 regularization, penalizes deviations from important parameters of previously learned tasks. By constraining significant weights to remain close to their prior values, the framework ensures that newly learned information does not overwrite essential knowledge. While round-based systems provide more predictable convergence patterns, FedCCL's asynchronous updates introduce non-determinism in the training process. 

\subsection{Privacy Preservation}
FedCCL ensures privacy by keeping raw data on client devices, sharing only model parameters and essential metadata through secure channels. The clustering mechanism adds privacy by grouping similar clients, making individual inference more difficult. These features enable deployment in sensitive domains like healthcare, finance, and industrial systems.

\subsection{Implementation}
The framework was implemented following production engineering practices with containerization, emphasizing scalability and security for real-world edge computing scenarios. All source code, datasets, and analysis scripts are publicly available\footnote{https://github.com/polaris-slo-cloud/fedccl}.

%% file: sections/case_study.tex
\section{Case Study: Energy Forecasting with FedCCL} \label{section:case_study}

Our case study implements and evaluates FedCCL using a real-world scenario for solar production forecasting of individual households and commercial \& industrial (C\&I) installations across central Europe. Such predictions are critical in the transition towards renewable energy systems, especially for use-cases such as battery charging, heating, and electric vehicle charging during periods of peak energy availability.

\begin{figure}[htbp]
\centerline{\includegraphics[width=0.5\textwidth]{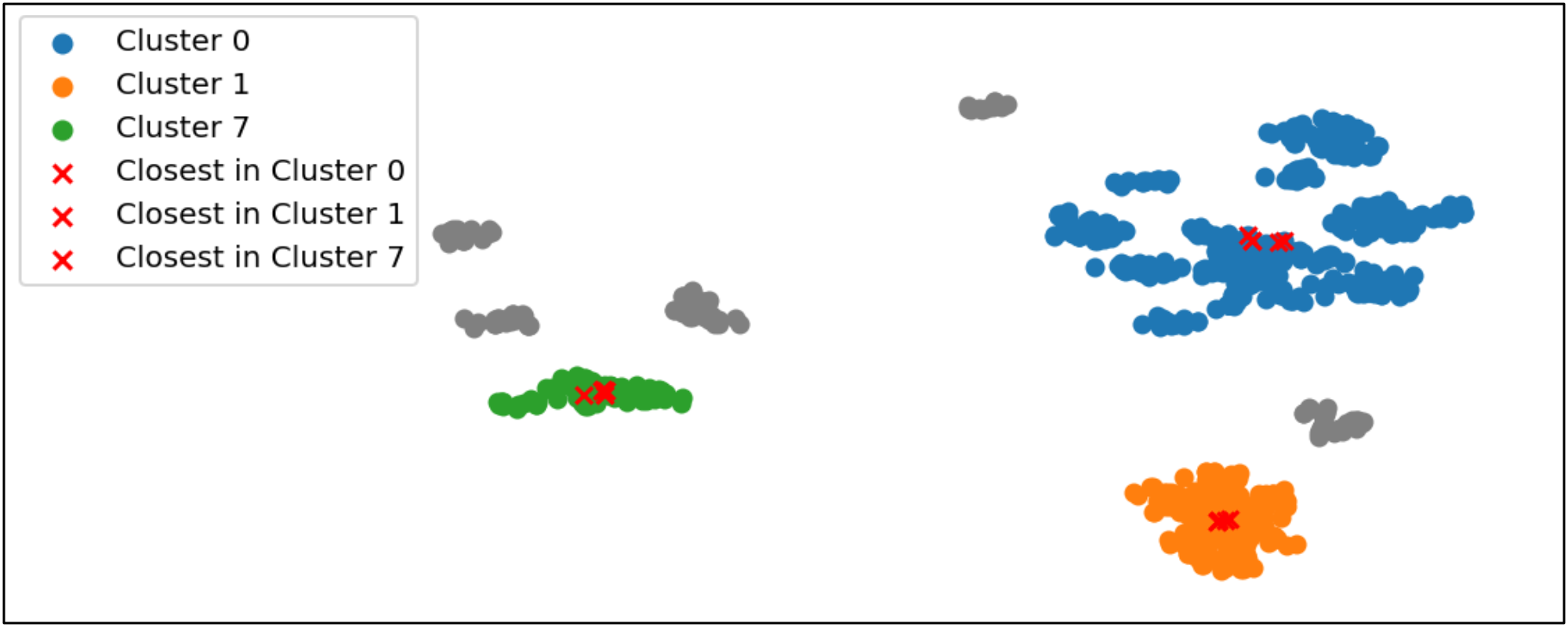}}
\caption{Location-based Clustering (Coordinates Hidden)}
\label{fig:site_clustering}
\end{figure}

The clustering phase in our study follows two primary objectives. First, we implement location-based clustering, as illustrated in Figure \ref{fig:site_clustering}, where installations are grouped based on their geographical proximity. 

This approach exploits the strong correlation between regional weather patterns and solar production. The figure demonstrates the distribution of three main clusters (marked in blue, orange, and green), with test sites (marked with red crosses) selected as the installations closest to each cluster's centroid. For privacy preservation, the exact longitude and latitude coordinates are intentionally obscured. Second, we incorporate installation-specific characteristics by clustering based on panel orientation, determined through analysis of peak power generation patterns during clear-sky conditions. This dual-clustering strategy ensures the framework can effectively account for both regional weather patterns and site-specific performance characteristics.

\subsection{Dataset and Forecasting Objective}
The dataset spans approximately fifteen months, comprising solar production data recorded at 15-minute intervals (96 measurements per day) and complementary hourly weather forecast data, which includes essential meteorological parameters affecting solar production. To harmonize these different temporal resolutions, weather forecasts are duplicated across their respective 15-minute intervals.

\begin{figure}[htbp]
\centerline{\includegraphics[width=0.5\textwidth]{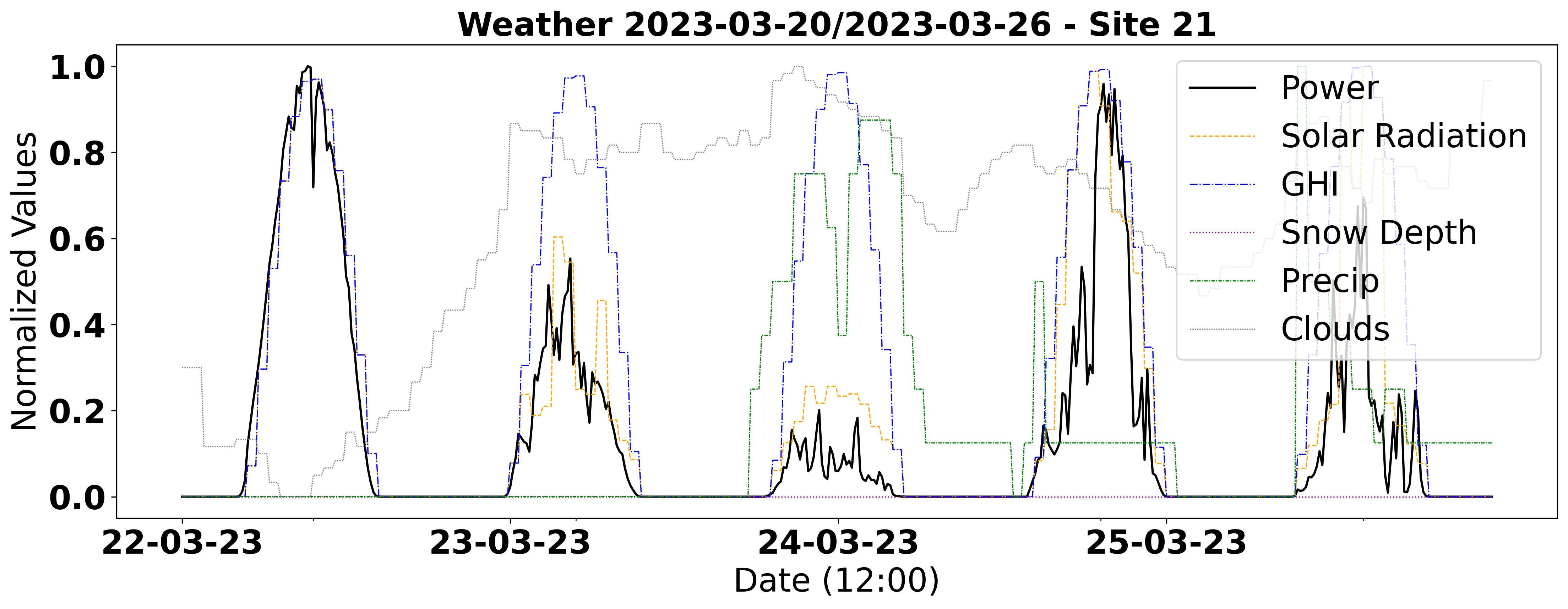}}
\caption{Data Shape}
\label{fig:data_shape}
\end{figure}

Figure \ref{fig:data_shape} illustrates a representative week of normalized data, demonstrating the relationship between meteorological conditions and power generation. The black line represents the actual power output. The daily production cycles are clearly visible, with peak generation typically occurring during midday hours when solar radiation is strongest. The influence of weather conditions is evident in the varying production patterns: days with high cloud coverage (purple) show reduced and more irregular power generation, while clear days exhibit smooth, bell-shaped production curves. Snow depth (green) and precipitation (red) remain at zero during this period. The training window for predictions encompasses seven days of historical data. Based on this historical data and weather forecasts, the model predicts power generation for the next 24 hours in 15-minute intervals, providing 96 discrete forecasting points per day.

\subsection{Feature Selection and Processing}
We identified key features that impact prediction accuracy. The most influential weather-related parameters include ground-level solar radiation and Global Horizontal Irradiance (GHI), which directly correlate with potential energy production. Snow depth and precipitation levels are incorporated to account for weather conditions that may obstruct panel efficiency, while cloud coverage data helps identify periods of potentially irregular production. Temporal features are used for capturing both daily and seasonal patterns in solar energy production. The system tracks the minute of day and day of year, both normalized to capture cyclical patterns in solar availability. All features undergo normalization to ensure consistent training across sites with varying capacities. Weather parameters are scaled using regional maximum values typical for central Europe, while production values are normalized relative to each site's maximum capacity (kWp).

\begin{table}[ht]
\caption{Feature Descriptions}
\centering
\begin{tabular}{|l|p{3.0cm}|c|}
\hline
\textbf{Feature Name} & \textbf{Description} & \textbf{Range (min. - max.)} \\ \hline
\textit{solar\_rad} & Ground-level solar radiation (W/m²) & 0 - 956.2 W/m² \\ \hline
\textit{ghi} & Global Horizontal Irradiance measured outside atmosphere (W/m²) & 0 - 956.21 W/m² \\ \hline
\textit{snow\_depth} & Snow depth affecting panel efficiency (mm) & 0 - 1178.6 mm \\ \hline
\textit{precip} & Precipitation levels (mm) & 0 - 14.78 mm \\ \hline
\textit{clouds} & Cloud coverage (high-coverage days filtered) & 0 - 100.0 \% \\ \hline
\textit{minute\_of\_day} & Temporal feature capturing daily cycles (minutes) & 0 - 1440 minutes \\ \hline
\textit{day\_of\_year} & Temporal feature capturing seasonal cycles (days) & 0 - 365 days \\ \hline
\end{tabular}
\label{tab:feature_table}
\end{table}

%% file: sections/evaluation.tex
\section{Evaluation} \label{section:evaluation}

\subsection{Evaluation Objective and Baselines}
We evaluate FedCCL against centralized baselines (CentralizedAll with complete data access and CentralizedContinual with progressive data availability) to demonstrate that we achieve similar or better performance and that clustering provides benefits over a globally shared model. Each client trains four models: local, federated global, and two federated cluster models (location-based and orientation-based). Using an 80-20 training-testing split across 100 experimental runs, we measure performance during both 24-hour periods and daytime operations (06:00-21:00) to ensure statistical significance and context-appropriate evaluation in real-world scenarios.

\subsection{Performance Metrics}
We assess prediction accuracy using two distinct Mean Absolute Percentage Error (MAPE) metrics. For instantaneous power predictions, we normalize errors against each system's rated capacity (kilowatts peak, kWp):
\[
\text{Power Error} = \left(\frac{\lvert\text{predicted} - \text{actual}\rvert}{\text{kWp}}\right) \times 100\%
\]
For cumulative energy we normalize errors against the theoretical daily maximum production, calculated as 12 hours at rated capacity:
\[
\text{Energy Error} = \left(\frac{\lvert\text{predicted} - \text{actual}\rvert}{\text{kWp} \times 12}\right) \times 100\%
\]

This normalization enables fair comparison across solar installations regardless of their size. We evaluate performance using mean and maximum values of \textit{error power} for instantaneous predictions and \textit{error energy} for cumulative predictions, allowing us to assess both typical performance and edge cases. Detailed results are presented in Table \ref{tab:model_comparison}.

\subsection{Baseline Performance Analysis}
The CentralizedAll model, with complete initial data access, establishes our primary baseline with a mean error for power prediction of 6.24\% (±0.07\%). This relatively low standard deviation indicates highly consistent performance across different runs. During daytime hours (6:00-21:00), the mean error increases to 9.27\%. The model achieves a mean error for energy prediction of 3.46\%, demonstrating better accuracy in cumulative energy prediction compared to instantaneous power forecasting.

The CentralizedContinual model, which mirrors real-world data availability constraints, shows degraded performance with a mean error power of 6.73\% (±0.14\%). This 0.49 percentage point increase in error compared to the CentralizedAll model quantifies the impact of progressive data availability on prediction accuracy. The higher standard deviation also indicates less consistent performance across runs.

\begin{table*}[!htbp]
\vspace*{-\baselineskip}
\centering
\small
\setlength{\tabcolsep}{4pt}
\renewcommand{\arraystretch}{1.3}
\caption{Comprehensive Model Performance Comparison (All values in percentages, based on 100 runs)}
\begin{tabular}{lcccccc}
\toprule
\textbf{Metric} & \textbf{Centralized} & \textbf{Centralized} & \textbf{Federated} & \textbf{Federated} & \textbf{Federated} & \textbf{Federated} \\
& \textbf{(All)} & \textbf{(Continual)} & \textbf{Global} & \textbf{Location} & \textbf{Orientation} & \textbf{Local} \\
\midrule
\multicolumn{7}{l}{\textbf{Overall Performance}} \\
Mean Error Power & 6.24 $\pm$ 0.07 & 6.73 $\pm$ 0.14 & 6.59 $\pm$ 0.24 & 6.44 $\pm$ 0.17 & 6.54 $\pm$ 0.09 & 6.34 $\pm$ 0.07 \\
Max Error Power & 61.60 $\pm$ 1.56 & 84.31 $\pm$ 1.10 & 70.34 $\pm$ 1.70 & 71.00 $\pm$ 1.70 & 70.80 $\pm$ 1.96 & 83.90 $\pm$ 1.06 \\
Mean Error Energy & 3.46 $\pm$ 0.08 & 4.18 $\pm$ 0.18 & 4.17 $\pm$ 0.44 & 3.93 $\pm$ 0.21 & 4.05 $\pm$ 0.19 & 4.19 $\pm$ 0.12 \\
\midrule
\multicolumn{7}{l}{\textbf{Daytime Performance (6:00-21:00)}} \\
Mean Error Day Power & 9.27 $\pm$ 0.11 & 10.01 $\pm$ 0.21 & 9.75 $\pm$ 0.36 & 9.56 $\pm$ 0.26 & 9.71 $\pm$ 0.13 & 9.41 $\pm$ 0.11 \\
Mean Error Day Energy & 4.38 $\pm$ 0.10 & 5.26 $\pm$ 0.21 & 5.25 $\pm$ 0.52 & 4.96 $\pm$ 0.25 & 5.16 $\pm$ 0.23 & 5.33 $\pm$ 0.15 \\
\bottomrule
\end{tabular}
\label{tab:model_comparison}
\end{table*}

\subsection{Federated Model Performance Analysis}

\subsubsection{Global Model}
The Federated Global model achieves a mean error power of 6.59\% (±0.24\%), positioning its performance between the two centralized baselines. The increased standard deviation suggests more variability in the accuracy of the prediction, likely due to the asynchronous nature of federated updates. Notably, the model's maximum error power of 70.34\% represents a significant improvement over the CentralizedContinual model's 84.31\%, indicating better handling of extreme cases.

\subsubsection{Location-Based Clustering}
\begin{figure}[htbp]
\centering
\includegraphics[width=0.5\textwidth]{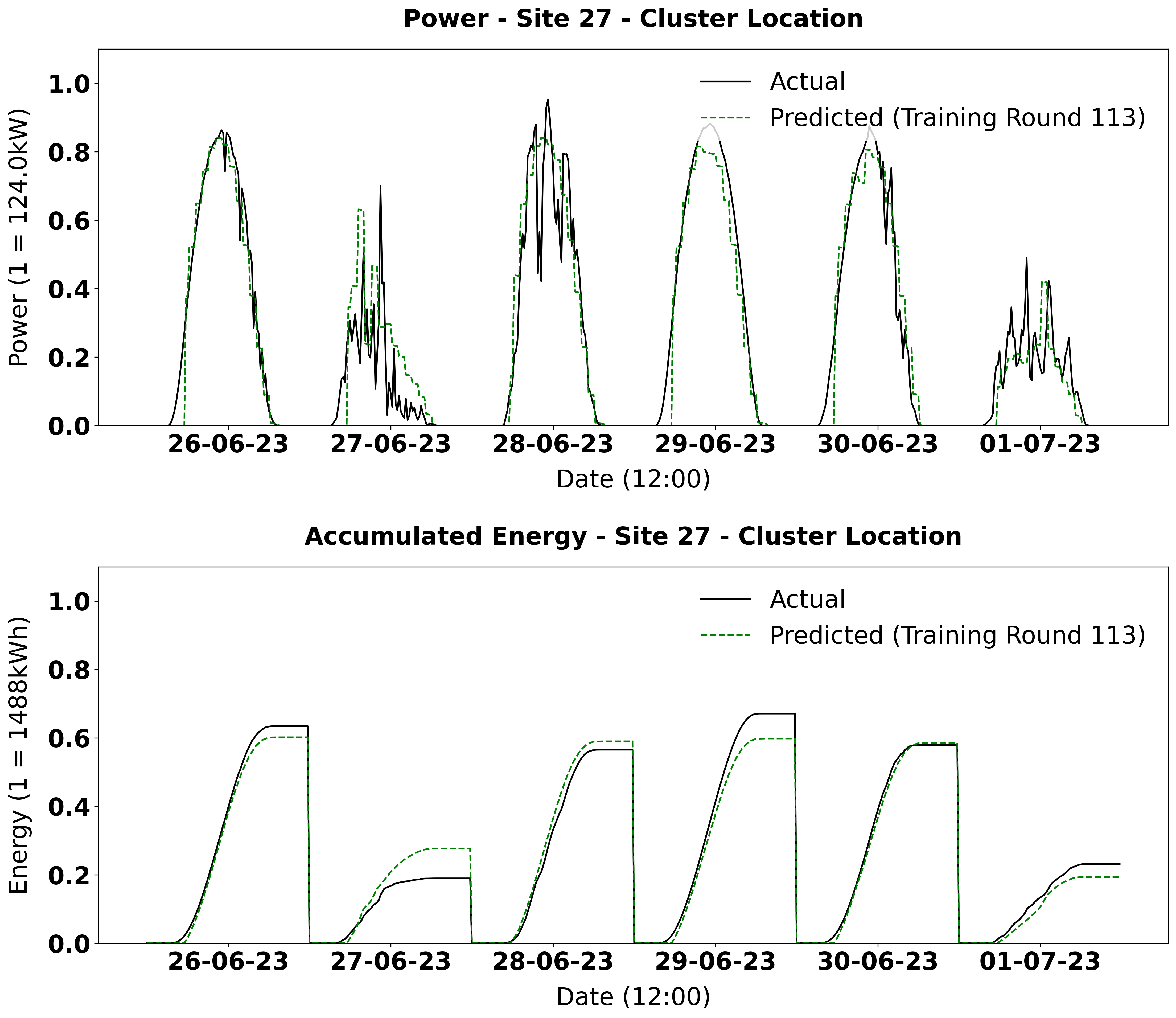}
\vspace{-10pt}
\caption{Location-based Cluster Model Performance Example}
\vspace{-10pt}
\label{fig:location_performance}
\end{figure}

The Location-based clustering approach demonstrates superior performance among federated models, achieving a mean error power of 6.44\% (±0.17\%). This represents only a 0.20 percentage point degradation compared to the CentralizedAll baseline while maintaining complete data privacy. The model's daytime performance (9.56\% mean error) shows particular strength in handling peak production periods, as illustrated by the exemplary prediction shown in Figure \ref{fig:location_performance}.

\subsubsection{Orientation-Based Clustering}
The Orientation-based clustering model shows slightly diminished performance with a mean error power of 6.54\% (±0.09\%). While this remains competitive, the smaller improvement over the Global model suggests that panel orientation alone may not be as strong a differentiator for model specialization as geographical location. However, this model demonstrates the highest consistency across runs, as indicated by its low standard deviation.

\subsection{Population Independent Performance Analysis}\label{section:population_independent}
\begin{figure}[htbp]
\centerline{\includegraphics[width=0.5\textwidth]{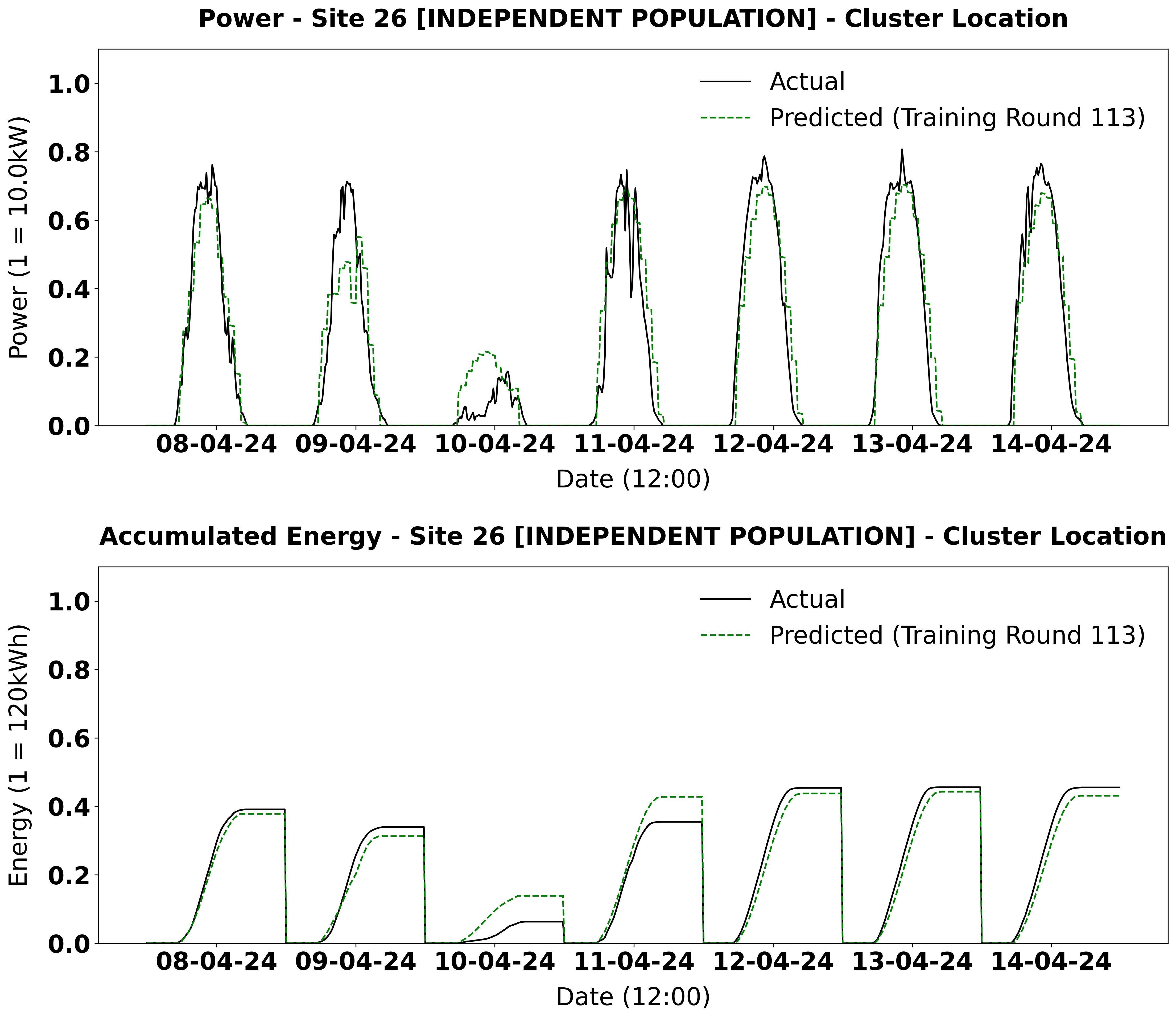}}
\vspace{-10pt}
\caption{Population Independent Performance Example}
\vspace{-10pt}
\label{fig:independent_performance}
\end{figure}

A key aspect of our evaluation, namely the FedCCL Predict \&Evolve contribution, focuses on the framework's ability to serve new installations without prior training data. This analysis reveals several key insights:

The Global and Location models demonstrate robust generalization capabilities, achieving mean error power rates of 6.60\% and 6.58\% respectively for independent sites. These results represent minimal degradation from their performance on the training population (0.01 and 0.14 percentage points), indicating strong transferability of learned patterns. The Orientation model shows less effective generalization with a mean error power of 9.13\% for independent sites, a 2.59 percentage point increase from its training population performance. Illustrated by the exemplary prediction shown in Figure \ref{fig:independent_performance}.

During daytime hours, the performance difference becomes more pronounced, with the Location model maintaining a mean error day power of 9.72\% compared to the Orientation model's 13.73\%. This divergence emphasizes the superior generalization capabilities of location-based clustering for new installations.

\subsection{Energy vs. Power Prediction Analysis}
Energy prediction consistently demonstrates lower error rates than power prediction across all models. The CentralizedAll model achieves 3.46\% mean energy error, while federated approaches range from 3.93\% to 4.19\%. Since energy is calculated through integration of power over time, this improved accuracy stems from an averaging effect that diminishes instantaneous prediction errors. This pattern persists in independent populations, with the Location model achieving 4.13\% mean energy error. This particularly benefits day-ahead planning and resource allocation applications.

\subsection{Implications for Real-World Deployment}
FedCCL provides a real-world applicable system delivering centralized-level performance while preserving privacy. The Location model (6.44\% error) outperforms the Global model (6.59\%), with minimal degradation ($<0.15$ percentage points) in new deployments, offering immediate value across various application domains without compromising privacy.

%% file: sections/related_work.tex
\section{Related Work} \label{section:related_work}
Centralized solar production forecasting has been extensively studied in the literature, with comprehensive reviews \cite{kumari_deep_2021, qing_hourly_2018, yu_deep_2024} documenting significant advances in prediction methodologies. Long Short-Term Memory (LSTM) networks \cite{hochreiter_long_1997}, which we also employ in our FedCCL Case Study, have established themselves as a leading approach for time series prediction in this domain. Their effectiveness has been validated across multiple studies \cite{jailani_investigating_2023, fungtammasan_convolutional_2023, abdel-nasser_accurate_2019}, consistently demonstrating robust performance in handling temporal dependencies inherent in solar production. Suresh et al. \cite{suresh_probabilistic_2022} achieved 2.6\% MAE relative to peak capacity for hour-ahead forecasting. While their shorter prediction horizon differs from our multi-day weather-based forecasts, our evaluation shows comparable error magnitudes. While centralized approaches have shown promising results, they face fundamental limitations in scenarios requiring data privacy. The emergence of privacy-preserving distributed learning in the energy domain has created three distinct categories of solutions, each addressing different aspects: traditional FL approaches, clustering-enhanced methods, and domain-specific optimizations.

\textit{Traditional Federated Learning Approaches.} Traditional FL implementations in energy forecasting, such as work from Stanford University \cite{meghana_bharadwaj_energy_2023} and STANN for GHI forecasting \cite{wen_solar_2022}, as well recent work exploring the other side, i.e., load forecasting. \cite{saputra_energy_2019, richter_advancing_2024} demonstrate the fundamental viability of FL in that domain. However, these approaches rely on synchronous training rounds and single global models, which create two significant limitations. First, the synchronous nature requires all participants to complete their updates before proceeding, creating potential bottlenecks when participants operate at different speeds or face connectivity issues. Second, the single global model struggles to capture local variations effectively. FedCCL addresses these limitations through its asynchronous architecture and multi-tiered model hierarchy.

\textit{Clustering-Enhanced Methods.} More sophisticated approaches have emerged that incorporate clustering to improve prediction accuracy. The FL+HC system for residential load forecasting \cite{christopher_briggs_federated_2022} demonstrate that grouped learning can improve model performance by 5\% or more. However, this method establish cluster after N rounds of training and clients belong to one cluster only. Similarly, fuzzy clustered FL for solar generation \cite{yoo_fuzzy_2022} also relies on cluster generation during training. Some approaches, like Tun et al.'s predefined clustering system \cite{tun_federated_2021}, move closer to FedCCL's philosophy but still rely on synchronous updates. FedCCL differs by performing clustering before training based on static problem specific characteristics and using asynchronous communication, better suited for real-world scenarios.

\textit{Domain-Specific Optimizations.} Recent work has focused on incorporating domain knowledge into FL architectures, particularly for solar forecasting. Physics-constrained approaches \cite{luo_deep_2021} and specialized deep learning architectures \cite{yu_deep_2024} demonstrate the importance of domain expertise. Edge computing implementations \cite{savi_short-term_2021} have shown promise in reducing communication overhead but typically maintain traditional FL patterns like a single global model and synchronous rounds. FedCCL builds upon these insights by combining domain-aware clustering (such as geographic location and panel orientation) with flexible client participation.

While recent approaches have explored individual elements - such as clustering for FL \cite{christopher_briggs_federated_2022,yoo_fuzzy_2022,tun_federated_2021}, domain-specific optimizations \cite{luo_deep_2021,savi_short-term_2021}, and basic privacy-preserving frameworks \cite{meghana_bharadwaj_energy_2023,wen_solar_2022} - no existing solution combines these aspects effectively. Current clustering methods require dynamic updates during training, domain-specific approaches remain limited by synchronous operations, and traditional implementations cannot capture local variations. FedCC uniquely integrates these components through pre-training clustering and asynchronous updates, achieving accuracy comparable to centralized approaches \cite{suresh_probabilistic_2022} while maintaining privacy and enabling immediate model specialization.

%% file: sections/conclusion.tex
\section{Conclusion \& Future Work} \label{section:conclusion}
This paper presents FedCCL, a framework that combines pre-training clustering with asynchronous FL. Through comprehensive evaluation in solar energy forecasting, FedCCL demonstrates performance that matches and exceeds centralized approaches while preserving data privacy and enabling immediate value delivery through pre-training clustering. The three main contributions — the FedCCL Framework, FedCCL Predict \& Evolve, and the FedCCL Case Study — establish both theoretical foundations and practical viability. The location-based clustering approach achieved 6.44\% mean error in power prediction, outperforming the CentralizedContinual baseline by 0.29 percentage points while maintaining data privacy. The framework's robust performance in population-independent scenarios further validates its usefulness for deployment to new installations. We intend to extend the FedCCL framework in multiple directions by investigating cluster granularity and intersection handling, specifically examining the impact of hierarchical sub-clusters on model performance and developing optimal strategies for managing overlapping cluster characteristics. Furthermore, defining definite criteria which model to use in the inference phase. Lastly, areas for exploration encompass framework generalization to other domains with inherent clustering properties and analysis of convergence characteristics. Given the framework's demonstrated effectiveness in solar forecasting, these research directions promise to advance the foundations of privacy-preserving distributed learning while enabling new industrial applications.

%% file: acknowledgements.tex
\section{Acknowledgements}
This work is partially funded by the Austrian Research Promotion Agency (FFG) as part of the RapidREC project (No. 903884). 
This research was funded by the EU’s Horizon Europe Research and Innovation Program as part of the NexaSphere project (GA No. 101192912).
This work has received funding from the Austrian Internet Stiftung under the NetIdee project LEOTrek (ID~7442).
The data has been generously provided by neoom AG.